# WHERE ARE THE HUMANS? A SCOPING REVIEW OF FAIRNESS IN MULTI-AGENT AI SYSTEMS

*Completed Research Paper*

Simeon Allmendinger, University of Bayreuth & Fraunhofer FIT, Bayreuth, Germany,
simeon.allmendinger@uni-bayreuth.de

Luca Deck, University of Bayreuth & Fraunhofer FIT, Bayreuth, Germany,
luca.deck@uni-bayreuth.de

Lucas Müller, University of Bayreuth, Bayreuth, Germany,
lucas.c.mueller@gmail.com

## Abstract

*Rapid advances in Generative AI are giving rise to increasingly sophisticated Multi-Agent AI (MAAI) systems. While AI fairness has been extensively studied in traditional predictive scenarios, its examination in MAAI remains nascent and fragmented. This scoping review critically synthesizes existing research on fairness in MAAI systems. Through a qualitative content analysis of 23 selected studies, we identify five archetypal approaches. Our findings reveal that fairness in MAAI systems is often addressed superficially, lacks robust normative foundations, and frequently overlooks the complex dynamics introduced by agent autonomy and system-level interactions. We argue that fairness must be embedded structurally throughout the development lifecycle of MAAI, rather than appended as a post-hoc consideration. Meaningful evaluation requires explicit human oversight, normative clarity, and a precise articulation of fairness objectives and beneficiaries. This review provides a foundation for advancing fairness research in MAAI systems by highlighting critical gaps, exposing prevailing limitations, and suggesting pathways.*

## 1 Introduction

Fueled by substantial investments in both human and physical capital, generative AI has witnessed remarkable progress in recent years (Maslej et al., 2024). Multi-Agent AI (MAAI) systems have emerged as a powerful paradigm for automating complex, knowledge-intensive tasks. They consist of multiple autonomous or semi-autonomous AI-based agents that collaboratively perceive, reason, and act within dynamic environments—structuring workflows, coordinating roles, and interacting with humans and systems through layered orchestration (Allmendinger et al., 2025). These systems have demonstrated impressive performance across diverse domains like software development (He et al., 2025) and financial analysis, e.g., Moody's employs MAAI systems in their day-to-day operations (Moody's, 2024). While fairness issues such as unjustified discrimination are well-documented in analytical AI systems (De-Arteaga et al., 2022), fairness research on generative AI, particularly Large Language Models (LLMs), is only starting to gain traction (Friedrich et al., 2025; Solaiman et al., 2025). Research on fairness in MAAI systems, specifically, is still in its infancy. The complexity, autonomy, and dynamics inherent to these systems introduce novel risks and unexpected behaviors (Chan et al., 2023; Deck et al., 2026). For example, when agents within a recruiting pipeline have differing fairness notions or goals, it is unclear how such conflicting interests are resolved and how the results correspond to overarching fairness desiderata such as non-discrimination. This prompts the research question:

*How does current research conceptualize, operationalize, and evaluate fairness in MAAI systems?*

To that end, we conduct a scoping review (Arksey & O'Malley, 2005) that synthesizes existing research and maps out the normative foundations as well as methodological approaches for operationalization and evaluation of fairness in MAAI systems. On this basis, we discuss how existing research contributes to established fairness goals and needs ("fairness desiderata") from interdisciplinary literature (Deck, Schoeffer, et al., 2024). This paper makes three key contributions to the emerging field of fairness in MAAI systems. First, it presents the first scoping review on fairness in MAAI, addressing a significant research gap by adopting a human-centric (Kühl, 2024), desiderata-driven (Deck, Schoeffer, et al., 2024), and system-level (Allmendinger et al., 2025) perspective—a departure from prior work focused predominantly on single-agent or model-centric fairness. Second, it conceptually systematizes the field by introducing a framework, through which it identifies five recurring fairness archetypes—*Normative Delegation, Fairness Facade, Fairness Schooling, Petri Dish Fairness, and Fairness Effectiveness*. In this emerging field, these archetypes capture early methodological and normative patterns across current research, which may help researchers identify prevailing assumptions, recognize underexplored topics, and position their own work within or against these patterns. Third, the paper offers a critical analysis of these archetypes in relation to established fairness desiderata, thereby revealing conceptual blind spots and methodological gaps, and providing an agenda for future research.

The paper begins with the three guiding perspectives, followed by foundational work on MAAI and fairness concepts. Afterwards, we outline the review approach and present five fairness archetypes as a result of the scoping review. Finally, we discuss our findings and suggest directions for future research.

## 2  Perspectives

To systematically map out the emerging field of fairness in MAAI, we adopt a particular perspective grounded in prior work. Specifically, we adopt a (1) human-centric, (2) desiderata-driven, and (3) system-level perspective. This approach allows us to move beyond purely formal algorithmic audits and instead capture fairness as a socio-technical property that emerges from the interaction of dynamic, heterogeneous agents. In the following, we describe these perspectives in detail.

**We investigate fairness in MAAI systems from a human-centric perspective.** The predominant focus of AI research has, for a long time, been on technical improvements concerning computational efficiency or performance metrics (Taylor et al., 2024). In contrast, human-centric AI "prioritizes designing, developing, and deploying AI systems that 'understand' human needs, enhance human performance and well-being, respect human rights, and align with human values" (Kühl, 2024, p. 6). Putting the needs and values of human stakeholders at the centre of AI development shapes the way in which ethical considerations such as fairness and transparency are considered (Schoeffer, 2022). As depicted in Figure 1, for an MAAI system to be fair in a human-centric sense always means to be either fair (i) from the perspective of or (ii) towards a certain human stakeholder.

**We expect AI systems to effectively fulfil fairness desiderata.** Fairness is among the most important ethical norms in both AI research (Binns, 2018; Mehrabi et al., 2021) and AI legislation (Deck, Müller, et al., 2024; European Commission & High Level Expert Group on Artificial Intelligence, 2019). A noteworthy characteristic of fairness is that it subsumes a range of desiderata (i.e., stakeholder needs an AI system is expected to satisfy) from multiple interdisciplinary perspectives (Deck, Schomäcker, et al., 2024; Mulligan et al., 2019) as well as several conflicting fairness objectives (Friedler et al., 2021; Mulligan et al., 2019). This has raised calls for effective mechanisms which require (i) concrete specification of fairness objective (Deck, Schomäcker, et al., 2024; Langer et al., 2024; Sterz et al., 2024) and (ii) rigorous evaluation of the claimed effects on fairness (Chen, 2023; Deck, Schoeffer, et al., 2024).

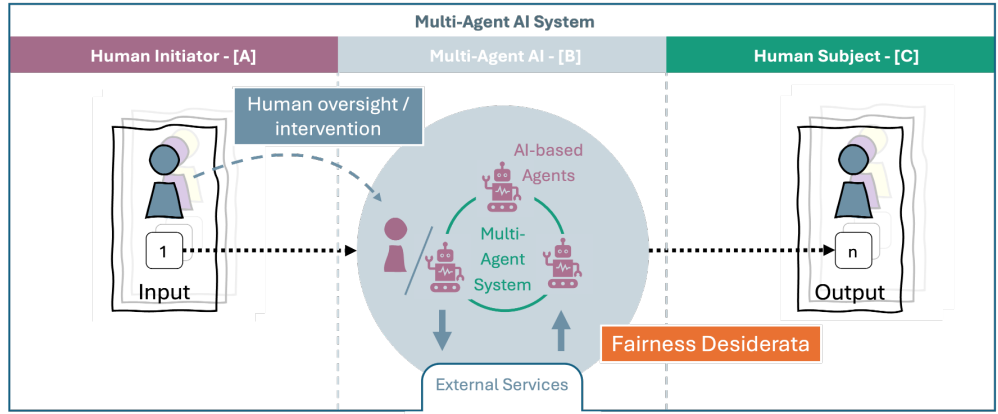

*Figure 1.*   *A Human-Centric Perspective on MAAI systems. This schema illustrates an AI-automated workflow with n tasks: A human initiator (A) submits an input request, which is processed by a Multi-Agent AI (B). This can include AI-based agents and human agents who may or may not be the original initiator. It can also contain interfaces to external data and services. The output is then delivered to a human subject (C), who also may or may not be the original initiator.*

Specifically, Deck, Schomäcker, et al. (2024) list eight fairness desiderata, which will be adopted for our scoping review: fairness understanding, data fairness, formal fairness, perceived fairness, fairness with human oversight, empowering fairness, long-term fairness, and informational fairness.

**We approach fairness in MAAI from a system-level rather than model-centric perspective.** While prior work often assesses fairness based on the outputs of a single foundation model (Friedrich et al., 2025) interacting with a human, our work assumes a broader perspective. Fairness in MAAI emerges—or fails—within the interactions of multiple AI-based agents. An MAAI system may produce unfair outcomes even if the underlying foundation models used by its AI-based agents behave fairly in isolation. This is because AI-based agent coordination, negotiation, and emergent dynamics introduce new fairness risks beyond probabilistic token generation. Hence, fairness in MAAI should not be conflated with fairness in LLMs alone but evaluated in light of systemic behaviors involving both AI-based agents and human stakeholders.

## 3   Foundations

Fairness in MAAI systems cannot be understood as a property of isolated models but must be conceptualized as a socio-technical outcome emerging from the interactions of multiple autonomous agents and human stakeholders. Accordingly, this section defines MAAI and repositions fairness as a normatively contested concept embedded in socio-technical contexts. It sheds light on a shift from static, model-centric metrics to human-centric, system-level frameworks that reflect the ethical complexity of real-world deployments.

**Multi-Agent AI systems.** MAAI systems are "system architectures composed of multiple autonomous or semi-autonomous AI-based agents that collaboratively perceive, reason, and act […] within shared, dynamic environments" (Allmendinger et al., 2025, p. 18). Grounded in foundation models such as LLMs, these agents demonstrate context-sensitive cognition, coordination, and adaptability—enabling the automation of complex, unstructured knowledge work beyond static rule-based systems. In contrast to traditional process automation, MAAI systems dynamically construct agent-integrated workflows, orchestrating roles and actions in real time to respond to evolving tasks and environments. This paradigm introduces novel capabilities but also raises critical challenges related to AI fairness, distributed autonomy, and the role of human oversight in socio-technical systems.

**AI Fairness.** With the increasing adoption of AI systems in high-stakes contexts, the last decade has sparked a tremendous amount of research on AI fairness in classification and risk scoring tasks such as credit scoring (Garcia et al., 2024), recruiting (Chen, 2023), or medical diagnosis (Barda et al., 2021).

AI fairness in these contexts is often understood as a property of AI systems that can be formalized and captured in fairness metrics. These metrics are manifold and can be based on statistical distributions, similarity measures, or causal reasoning (Verma & Rubin, 2018). Normatively speaking, fairness metrics typically assume some form of egalitarianism in the sense that "human beings are in some fundamental sense equal and that efforts should be made to avoid and correct certain forms of inequality" (Binns, 2018). Recently, AI fairness research is shifting from purely formal considerations to broader fairness desiderata (Deck, Schomäcker, et al., 2024; Mulligan et al., 2019). These include, among others, data fairness (i.e., properties of the training data such as representativeness (Buolamwini & Gebru, 2018)), perceived fairness (i.e., opinions and attitudes of affected individuals (Starke et al., 2022)), or informational fairness (i.e., transparency about how fairness is addressed (Schmude et al., 2025)).

**Fairness versus Ethics.** AI fairness is strongly related to AI ethics in general and inspired by a range of prominent theories from moral and political philosophy, such as egalitarianism (Binns, 2018), consequentialism (Card & Smith, 2020), or Rawls' justice as fairness (Franke, 2021). Though, while AI ethics is concerned with questions on the relationship between ethical norms and AI in general (Rességuier & Rodrigues, 2020), the field of AI fairness focuses on a specific set of norms (such as egalitarianism or consequentialism). Due to the novelty of the research gap, in this work, we are extending the scope of fairness towards ethical considerations in MAAI in general and discuss how these ethical considerations relate to fairness afterwards.

**Fairness in Generative AI.** However, fairness research on generative AI is still scarce. Friedrich et al. (2025) identify three patterns of fairness in generative AI: data pre-processing to remove undesired patterns before model training (e.g., by documenting (Yang et al., 2020) or filtering (Nichol et al., 2022) datasets), fairness constraints within the training algorithm (e.g., by adversarial learning (B. H. Zhang et al., 2018) or reinforcement learning (Y. Zhang et al., 2025)), and post-processing approaches to rectify undesired outputs after the model has been deployed (e.g., based on human instructions(Friedrich et al., 2025)). As previous research has primarily focused on singular models and formal aspects of fairness, such as feature distributions and metrics, this leaves a research gap for broader considerations of fairness in MAAI.

**Fairness in MAAI systems.** While foundational work has established the technical and normative complexities of fairness in AI, its specific articulation within MAAI systems remains fragmented and insufficiently examined. This gap is especially critical given the emergent dynamics, distributed autonomy, and shifting responsibilities across AI-based agents and stakeholders in shared environments of MAAI systems. Building on the conceptual foundations and the three analytical perspectives—human-centricity, fairness desiderata, and system-level framing—this work addresses the central research question: How does current research contribute to fairness desiderata in MAAI systems? The subsequent chapters synthesize existing literature to uncover how fairness is defined, embedded, and assessed in MAAI, thereby advancing a structured and normatively grounded understanding of this emerging field.

## 4 Method

Since fairness in MAAI is an emerging and fast-moving field, there is currently a lack of peer-reviewed papers. To still comprehensively capture early trends and synthesize research activities in such cases, scoping reviews provide a proven method to rigorously conduct and document the research process (Paré et al., 2015). Following the guidelines from Arksey & O'Malley (2005), our method comprises five stages: identifying the research question, identifying relevant papers, paper selection (Section 4.1), charting the data (Section 4.2), and collating, summarizing and reporting the results (Section 5).

### 4.1 Literature Review

Following our central research question and based on the methodology outlined by Kitchenham et al. (2007), we systematically identify existing research on fairness and ethics in MAAI systems. In total, we identified 533 potentially relevant records, as depicted in Figure 2. An initial exploratory review of the literature was conducted prior to finalizing the search strategy, in order to gain insights into the domain, assess keyword effectiveness, and identify relevant publishers. Our search string was designed to capture various components of MAAI systems while ensuring relevance to fairness-related contexts. The search string consisted of three parts, which are connected by the "AND" operator: *fairness, multi agent,* and *Large Language Model* based AI. For fairness and ethics, we used the terms "fair*" and "ethic*". As fairness in MAAI is a relatively young topic, we included ethics as an umbrella term that is commonly used synonymously or strongly related to fairness in MAAI research, e.g., Xu et al. (2025). The multi-agent aspects were covered with different forms of the term with the extension "agentic". For MAAI systems, we included related terms such as "Foundation Model*", "Large Language Model*", and "AI". To account for different word forms, including adjectives and nouns, we made extensive use of the asterisk, which includes words of the same stem, such as adjectives and nouns:

> ("fair*" OR "ethic*") AND ("multi-agent" OR "multi agent" OR "multiagent" OR "Agentic*") AND ("Large Language Model*" OR "LLM*" OR "*GPT" OR "Transformer*" OR "Foundation Model*" OR "AI" OR "Multimodal*" OR "Large Action Model*")

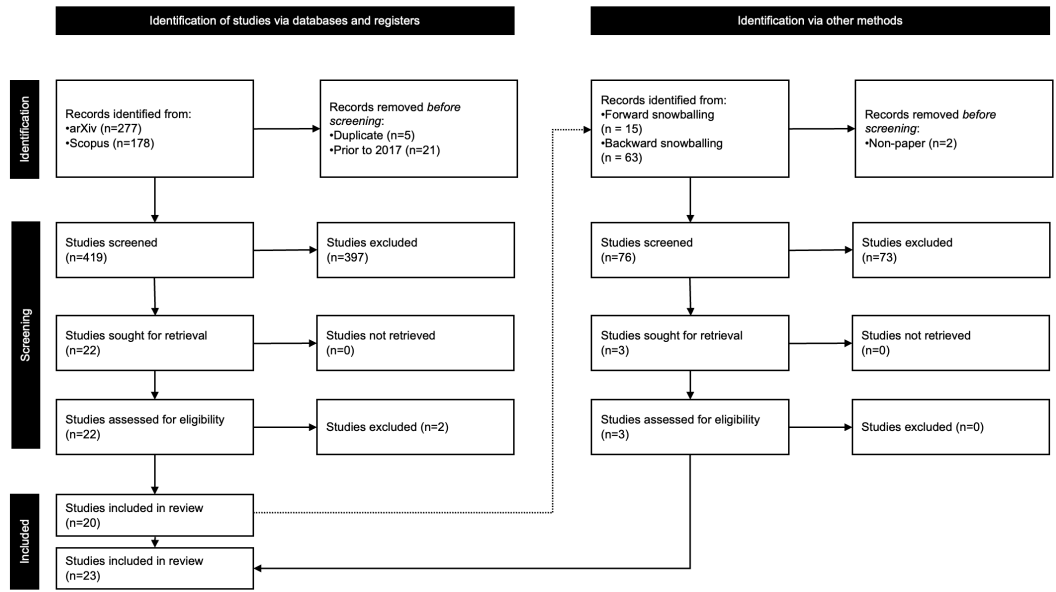

*Figure 2.      PRISMA flowchart illustrating the literature review process.*

Following proven guidelines for identification of primary studies (Wohlin et al., 2022), the Scopus database was chosen, due to its coverage of all relevant publishers, and augmented with forward and backward snowballing. To ensure the inclusion of recent, unpublished research, we also applied our search string to the arXiv database. In line with documentation standards (Kitchenham et al., 2007) and the PRISMA framework (Page et al., 2021), we ensured a transparent and reproducible selection process. Figure 2 visualizes how the initial set of 455 records (as of January 2025) was narrowed down to 20. After that, a backward and forward search was conducted, identifying 78 entries in total. Out of these, 3 met the inclusion criteria, resulting in a final set of 23 records. On the screening level, we manually reviewed each paper based on its abstract. To be included in this review, a paper had to address fairness norms in the context of an LLM-based MAAI system. We limited this review to LLM-based MAAI systems due to their pivotal role in advancing MAAI systems across a variety of domains, thereby making them an integral part of sophisticated MAAI systems (X. Li et al., 2024). Papers published from 2017 onward are considered, as the introduction of the transformer architecture in that year set the technical foundation for LLMs that constitute the systems under review (Vaswani et al., 2017). Out of the 23 papers, 20 were originally published on the arXiv database and 3 from Scopus. One study was published in 2023, 20 in 2024, and 2 in 2025. As of March 2026, eight of the 20 arXiv papers have since appeared in peer-reviewed venues, three as posters at NeurIPS, one as a poster at ICLR, and four in peer-reviewed journals. The high share of recently published and non-peer-reviewed papers highlights the novelty and growing interest in fairness in MAAI systems.

## 4.2   Charting the Data

To further structure and analyze the selected literature, we developed a *morphological box* that synthesizes key dimensions observed across the 23 final papers (see Figure 3). Drawing on principles of morphological analysis rooted in the works of (Zwicky, 1969), the dimensions of the morphological box were derived through a qualitative comparative conceptual analysis of the 23 reviewed papers. By examining each study in terms of how fairness is defined, how it is operationalized, and in what system context it is addressed, we identified five relevant axes for characterizing how fairness is addressed in MAAI systems: *Fairness Concept, Method, Fairness Inception, Objective,* and *Autonomy*. Fairness Concept distinguishes whether a paper grounds its understanding of fairness in normative theory (normatively grounded) or treats fairness as an intuitive, underspecified notion (simplistic). The Method dimension classifies the evaluation approach employed, spanning Output Analysis, Question \& Answers, Formal Games, Human Evaluation, and cases where the evaluation procedure remains Unclear. Fairness Inception captures the stage and mechanism by which fairness is introduced, ranging from post-deployment approaches such as prompting (Request) and role assignment (Multiple Role Assignments) to development-stage interventions through foundation model training (Teach) or fine-tuning (Educate). Because papers may simultaneously document, examine, and improve fairness outcomes, Objective is the only non-mutually exclusive dimension. Autonomy captures the agent composition of the evaluated system: solely AI-based configurations, hybrid human-AI settings, or purely human multi-agent arrangements. Figure 3 visualizes the distribution of papers across these dimensions. For example, most papers rely on simplistic fairness concepts and conduct lab experiments, while field deployments and educational approaches remain underrepresented.

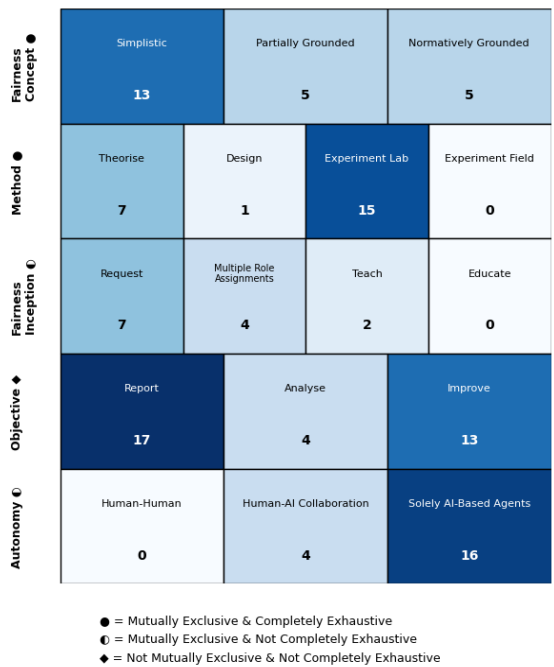

*Figure 3.   Overview of existing literature classified using a morphological box along five dimensions. A darker color indicates more occurrences of certain characteristics.*

Similarly, the majority of studies focus on settings involving solely AI-based agents, with limited attention to solely humans or hybrid autonomy in Human-AI Collaboration (HAIC). Expressions for which no papers were identified were deliberately retained to preserve the completeness of the conceptual space and to make research gaps explicit. To identify recurring patterns in how fairness is conceptualized and operationalized, we analyzed co-occurrences and relationships between dimensions within the morphological box. Specifically, we investigated how frequently papers appear in particular combinations across different dimensions (e.g., simplistic fairness in conjunction with the objective to improve fairness). These cross-dimensional patterns revealed meaningful tendencies in the literature. Based on these structural tendencies and the interpretive patterns that emerged from the analysis, notably around how fairness is grounded, introduced, and situated relative to human stakeholders, we derived a set of archetypes that characterize typical configurations of fairness-related research in MAAI. These archetypes synthesize both the structural tendencies and conceptual implications observed across the reviewed studies (see Figure 4).

# 5 Results: Archetypes of Fairness in MAAI

As MAAI systems increasingly mediate decision-making in sensitive domains, questions of fairness and normative responsibility gain heightened importance. Each of our five archetypes—representing different configurations of human and AI agency—implies distinct ways in which normative choices can be made, shared, or delegated. Yet, across the landscape of AI fairness research, we observe a recurring challenge: while fairness is frequently cited as a purported goal, it is often introduced in a superficial manner—grounded in intuitive notions and lacking engagement with normative theory. This issue becomes particularly salient when examining how normative responsibilities are handled or neglected within each archetype.

A large portion of the included papers attempts to improve fairness in MAAI systems without thoroughly introducing or motivating the type of fairness that is investigated. We refer to these fairness concepts as "simplistic" (simplistic → improve, 7 out of 13 cases). For example, Mushtaq et al. (2025) assign one out of ten agents in their MAAI system to evaluate engineering design proposals from societal, environmental, and ethical perspectives, without specifying how these concepts are understood or operationalized. In contrast, the few studies that are normatively grounded show a conceptual foundation (normatively grounded → improve, 3 out of 5 cases), such as Piatti et al. (2024) who instruct their agents to act in line with the categorical imperative of Kant (1998). Generally, the papers vary significantly in their use of the terms fairness and ethics. Out of 23 papers, 7 refer to ethics, 5 to fairness, and 11 to both. Papers using both "fairness" and "ethics" mostly do not treat them as synonyms—with exceptions such as Xu et al. (2025) and Zhao et al. (2024). Occasionally, ethical considerations extend beyond fairness, addressing implementation issues or broader ethical concerns like security risks (e.g., jailbreaking (Gan et al., 2024)). However, of the 18 papers that address ethics—7 focusing exclusively on ethics while 11 also refer to fairness—a majority do not clarify how they define the term or which conception they apply. For instance, Mushtaq et al. (2025) instruct one of their AI-based agents to act as an expert in ethics without specifying which ethical conception to follow. This marks a shift in responsibility by transferring the obligation to ensure fairness from human oversight to an AI-based agent within the MAAI system. Notably, the normative foundation of AI-based agents is often unclear and is developed in external development spheres such as training of foundation models (see Figure 5). Therefore, we refer to this archetype as *Normative Delegation*. A further archetype that emerges from our analysis is *Petri Dish Fairness*, which reflects the absence or underrepresentation of human involvement in the operation and evaluation of fairness within MAAI systems. A substantial number of studies are conducted in controlled lab environments and focus exclusively on interactions between AI-based agents (experiment lab → solely AI-based agents, 12 out of 15 cases). In these isolated configurations, fairness considerations are confined to AI-based agent dynamics, leaving open critical questions about the meaning and applicability of fairness in the absence of human stakeholders. This pattern is especially pronounced in formal game studies such as the Ultimatum Game (Lin et al., 2024; Noh & Chang, 2024), where agents often behave irrationally or contradict fairness intuitions (e.g., irrational rejections or failure to exploit monopolies). The absence of human participation stands in contrast to the relatively rare cases of HAIC, where fairness is explored in hybrid systems involving both human and AI-based agents (HAIC, 4 out of 23 cases). Of these, three studies carry out experiments, whereas one remains theoretical. This significant imbalance suggests that fairness is often decoupled from the social contexts in which it ultimately matters. The archetype *Fairness Effectiveness* reflects papers that seek to improve fairness based on benchmarks, structured comparisons, or experimental interventions that yield measurable improvements. This serves two purposes. First, implementing quantifiable metrics transforms fairness from an abstract, high-level concept into a concrete, measurable objective embedded in the implementation process. This increases transparency, comparability, and reproducibility. Second, it provides a normative basis for a debate on the choice and implementation of fairness objectives. Therefore, these contributions mark an important shift toward evidence-based approaches for evaluating and improving fairness. We observe that 17 out of 23 papers explicitly report fairness or ethical performance utilizing a large variety in methodology including *Output Analysis* (3), *Question & Answers (Q&A)* (5), *Formal Games* (5), *Human Evaluation* (2), and *Unclear Evaluations* (2).

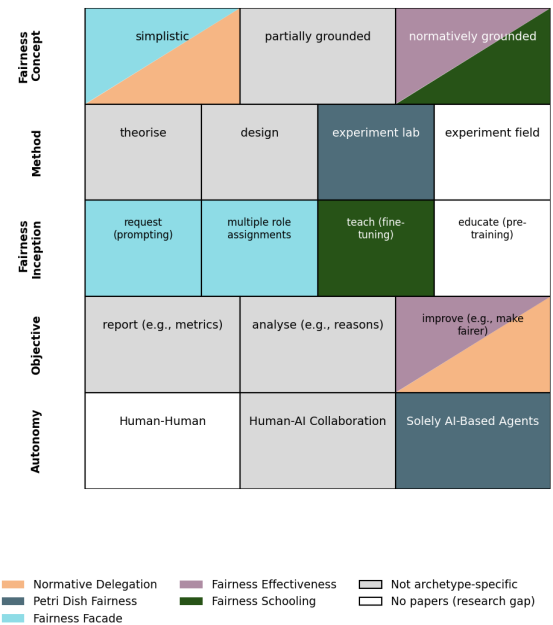

*Figure 4. Archetypes categorized via the morphological box. Shared cells (multiple colors) indicate papers exhibiting characteristics of more than one value within dimension. Gray cells indicate papers that do not define a specific archetype, while white cells highlight research gaps.*

Output Analysis involves the use of a variety of techniques to evaluate the textual outputs of agents. For example, (Cerqueira et al., 2024) utilize GPT-4o to compare the output of a single agent and an MAAI using thematic analysis. Combined with their hierarchical clustering and further analysis, they conclude that their MAAI system is more trustworthy and therefore more aligned with ethical principles, making it superior to the single-agent system. Central to the argument of Cerqueira et al. (2024) is the longer output of the MAAI system and corresponding increased occurrence and diversity of terms related to ethics. However, this approach presumes a direct and reliable causal relationship between purporting and actually carrying out ethical behavior—a presumption that remains tenuous and empirically unsupported. Q&A employs sets of ethical questions with predefined correct answers to evaluate the system based on accuracy. For example, Becker et al. (2024) uses the "Simple Ethical Questions" benchmark (Sitelew et al. 2021), in which agents must decide if historical personalities such as Hitler, Stalin, or Einstein are appropriate role models for children. This simplistic approach highlights the inadequacy of binary decisions in capturing the nuanced and complex nature of ethical judgments prevalent in real- world scenarios. Formal Games, derived from game theory, place agents within structured interactions. For example, Piatti et al. (2024) employ a "tragedy of the commons" scenario, assessing fairness via the Gini coefficient to measure equality of resource extraction among agents. This implicitly assumes an egalitarian normative framework, underscoring the often-implicit nature of fairness assumptions inherent in these studies. Human Evaluation involves human judgment to assess the alignment with fairness or ethical norms. Mushtaq et al. (2025), e.g., compare human assessments and MAAI-based evaluations of student-generated project proposals incorporating ethical considerations. Importantly, human evaluators possess the unique capability to capture the nuanced and implicit dimensions of fairness, thereby potentially enhancing the expressiveness and relevance of the evaluation. Unclear evaluation pertains to studies lacking transparency or reproducibility in their fairness evaluation methods. For example, (Bai et al., 2024) introduces a benchmark to assess fairness, revealing subtle patterns of discrimination—such as assigning technical or leadership roles based on perceived gender or ethnicity—that are absent in traditional single-agent systems. Although they describe the rationale behind their fairness benchmark, they fail to specify the concrete implementation or to provide accessible datasets, impeding reproducibility and adoption.

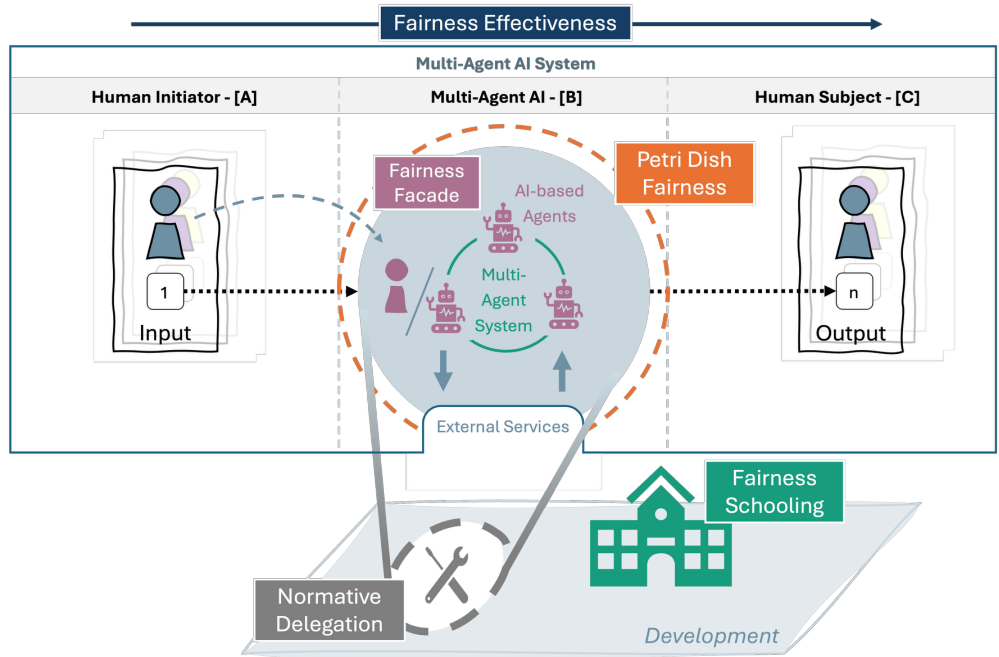

*Figure 5.    Extracted archetypes addressing the fairness desiderata of MAAI systems.*

We further identify the archetype *Fairness Facade*, characterized by the tendency to introduce fairness through superficial means such as high-level prompting or role assignment (simplistic → request, 5 out of 13 cases; simplistic → multiple role assignments, 2 out of 13 cases). In these cases, fairness is treated as an add-on rather than an integrated design principle. For example, several studies assign agent personalities or role definitions (e.g., "agreeable" or "cooperative") to promote desired negotiation outcomes, e.g., Noh & Chang (2024). Others attempt to achieve ethical behavior through generic prompts without grounding these prompts in structured learning processes, e.g., Piatti et al. (2024). Such techniques can substantially affect the agents' behavior in short-term interactions (Noh & Chang, 2024). However, they frequently lack robust normative foundations and do not ensure consistency across varying contexts.

Opposed to Fairness Facade, *Fairness Schooling* involves deliberate fairness instructions during MAAI development rather than prompts after deployment. This approach is rarely observed in the literature, with only one study (Tennant et al., 2024) normatively grounding its approach and explicitly aiming to teach fairness to agents with fine-tuning (normatively grounded → teach, 1 out of 5; all others → Teach, 1 out of 15). Despite this limited evidence, Fairness Schooling may present a nascent but promising shift—positioning fairness as a central design principle.

## 6    Discussion

In this section, we reflect on the key archetypes identified in our review, connecting them to established goals and requirements from prior literature. For this, we rely on Deck, Schomäcker et al. (2024), who distill eight "*fairness desiderata*" from interdisciplinary literature and map them to the stages of an AI lifecycle. We critically assess how current approaches to fairness in MAAI systems succeed or fall short with respect to these desiderata and outline promising directions and open challenges for future research.

### 6.1    What is the fairness based on?

According to Deck, Schomäcker et al. (2024), all fairness endeavors initially require an appropriate *Fairness Understanding*, including specification of concrete fairness objectives and normative reasoning about the selection of fairness objectives in a given context. However, the archetype *Normative Delegation* reveals that the definition of and responsibility for fairness is often delegated to AI-based agents and hence externalized to the MAAI development. This approach disregards the multidimensionality and contradictory nature of fairness, given that fairness spans across the entire AI lifecycle (e.g., data selection, preprocessing, model design, etc.) and across multiple desiderata (Deck, Schomäcker, et al., 2024). For example, highly unequal salary distributions may be considered fair from a libertarian perspective, e.g., Nozick (1974), if they arise through voluntary exchange. In contrast, egalitarian frameworks, e.g., Cohen (1979), would regard such disparities as inherently unjust. Moreover, fairness is often equated to an absence of "bias" (Bai et al., 2024), where bias can ambiguously mean a deviation from a true value in a statistical sense or deviation from a societally desirable value in a normative sense.

By contrast, the archetype *Fairness Effectiveness* shows a more promising alignment with *Fairness Understanding* by motivating and specifying fairness measures in a way that can be adopted and optimized in practice and future studies. For example, Mushtaq et al. (2025) operationalize their fairness conception through a benchmark, which they then use as the foundation to develop an MAAI system that out- performs competing approaches when measured against this benchmark. However, even in these cases, normative foundations often remain flat, and the focus is usually more on fairness outcomes rather than justifications. To fulfill the desideratum of fairness understanding, future work should consequently start with clarifications of the fairness objectives and underlying normative assumptions.

## 6.2 How to incept fairness in MAAI systems?

The archetype of Fairness Schooling is not only closely tied to *Fairness Understanding* but also to the desideratum *Formal Fairness*, which includes mathematical and statistical measures that can be embedded and evaluated during model development (Deck, Schomäcker, et al., 2024). Currently, the schooling of fairness—that is, teaching fairness norms during the development stages of MAAI systems (e.g., fine-tuning)—is largely absent. Only one reviewed study (Tennant et al., 2024) explicitly incorporates a normatively grounded concept of fairness into the creation of an AI-based agent by finetuning a lightweight LLM. However, *Fairness Schooling* requires accessible LLMs, which is not the case for closed-sourced models, such as the market-leading products of OpenAI, Google, and Anthropic (Xiao et al., 2024).

In contrast, the archetypes of *Normative Delegation* and *Fairness Facade* demonstrate less resource-intensive approaches. These papers incept fairness at the deployment stage, mostly relying on prompting or rule-based framing. This post-development approach fails to meet both desiderata. Neither does it foster a principled *Fairness Understanding* nor does it embed fairness objectives in training and evaluation environments, accounting for Formal Fairness. This disconnect highlights a broader issue: research on fairness in MAAI systems is currently in a state of convenient minimalism, doing solely what is easy and efficient to implement rather than what is conceptually or ethically sound. Furthermore, researchers should aim at developing artifacts that are reproducible and applicable in research and industry. As long as research is merely regarding fairness as an add-on feature rather than a foundation, fairness in MAAI systems will remain superficial. Overcoming this requires a paradigm shift from *Fairness Facade* as a fig leaf to *Fairness Schooling* as an integrated MAAI system development.

## 6.3 Does the inception of fairness decrease task performance?

In analytical AI, tradeoffs between *Formal Fairness* and task performance are well documented (J. Li & Li, 2025). Whether such tradeoffs exist in MAAI systems remains an open question, given the inconclusive evidence. Proponents of the thesis that fairness measures decrease task performance occur more often in the archetype *Fairness Facade*. For instance, Cheng et al. (2024, p. 4) warn that "[b]lindly applying fairness constraints [...] may greatly compromise the overall effectiveness of the solution method." In contrast, most studies of the archetype *Fairness Effectiveness* show that fairness measures can actually increase task performance, particularly in formal game settings. For example, Tennant et al. (2024) demonstrate that agents whose LLMs have been finetuned with either a deontological or a utilitarian reward function achieve higher payoffs in an iterative prisoner's dilemma game compared to those with a reward function geared towards payoff maximization. In a similar vein, Piatti et al. (2024) analyze a "tragedy of the commons" game, where AI-based agents must balance individual short-term against long-term rewards. When agents adhere to deontological principles, overall payoffs increase because they prioritize sustainable cooperation over immediate benefits. However, when a non-deontological egoistic AI-based agent is introduced into the system, it exploits the cooperative population, destabilizes the equilibrium, and secures a higher individual payoff at the expense of others (Piatti et al., 2024). This pattern of exploitation extends to other formal game scenarios. In Noh & Chang (2024), agents play an ultimatum game with varying personality assignments. AI agents making fairness-based arguments are more likely to reach agreements, leading to higher collective payoffs, but also more susceptible to exploitation by self-interested agents, indicating deviation from purely payoff-maximizing behavior. Though these findings originate only from controlled simulations, they indicate that fairness and performance are not inherently at odds, but rather that their relationship is context-dependent and influenced by MAAI system dynamics. Further research is required—particularly beyond petri dish experiments—to understand how fairness inception impacts various performance measures (e.g., *individual vs. collective and short-term vs. long-term payoffs) in* real-world applications.

### 6.4 How does the design of AI-based agents affect fairness?

Foundation models, such as LLMs, are the building blocks of AI-based agents in MAAI systems. Consequently, the choice of LLMs has implications for all desiderata outlined (Deck, Schomäcker, et al., 2024). With twelve instances, the GPT-4 family is the dominant group of LLMs in the surveyed papers and particularly prevalent among the archetypes *Normative Delegation* and *Fairness Facade*. The strong focus on GPT-4o improves comparability across studies and informs practitioners, given the substantial market share of OpenAI's models (Xiao et al., 2024). However, such a strong focus also carries the risk of overlooking other LLMs and underestimating important differences among models. Embedded political opinions, e.g., vary significantly across LLMs, highlighting the risks of depending on a single model (Ball et al., 2025). Furthermore, research comparing different LLMs indicates a positive correlation between intelligence and normative performance (Bai et al., 2024; Piatti et al., 2024). In addition, all reviewed studies creating MAAI systems utilize exclusively English inputs. However, Leng (2024) show that cognitive biases in LLMs can vary significantly with input language. For GPT4-Turbo, they find irrational loss aversion similar to humans when prompted in Spanish or French—but not in English. As MAAI systems are expected to operate globally with a variety of languages, further research on *Fairness Effectiveness* as well as *Fairness Schooling* is needed.

### 6.5 Which fairness dynamics emerge in MAAI systems?

Systems employing simple methods to incept fairness, characterized by the archetype Fairness Facade, consequently become more vulnerable to risks such as fairness drifts. This behavior relates to the desideratum Long-term Fairness, which emphasizes the downstream impact of deployed models and appropriate countermeasures (Deck, Schomäcker, et al., 2024). As an example, Lin et al. (2024) find that AI-based agents can collude in a response to the incentive structure, without any explicit instruction or communication. This leads to a loss of collective welfare within their formal experimental setup. Furthermore, AI-based agents can exhibit herding behavior, converging upon nonsensical and unethical positions if those represent the majority view (Cisneros-Velarde, 2024). If this behavior translates to other MAAI system contexts, unfair norms may be manifested and amplified once they are introduced to a system. The dynamics within MAAI systems are also influenced by the agents' underlying foundation models, as illustrated (Piatti et al., 2024). How MAAI systems, composed of different foundation models, affect fairness dynamics remains unclear, as no study has examined such a configuration. Future research should address this research gap by varying foundation models and composition.

### 6.6 Where are the humans?

Finally and most importantly, the archetype Petri Dish Fairness, characterized by the absence of human stakeholders in MAAI research, stands in direct contrast to two fairness desiderata outlined (Deck, Schomäcker, et al., 2024): *Fairness with Human Oversight* and *Perceived Fairness*. The former emphasizes the role of human oversight in AI decision-making processes with respect to concrete fairness objectives. However, our findings show that in the majority of papers, humans are only involved in the role of system designers or leaders of experiments, but not as an integral part of the MAAI system. Without a morally responsible human-in-the-loop, however, the studied MAAI systems lack practical applicability and do not allow any insights about *Fairness with Human Oversight*.

Similarly, the desideratum of *Perceived Fairness*, i.e., accounting for fairness perceptions of affected individuals, is undermined when MAAI systems only include AI-based agents in "petri-dish"-like settings. When there is no human to perceive, evaluate, or interact with the fairness of processes and outcomes, fairness in MAAI remains conceptually and practically detached from the very purpose human-centric AI is supposed to fulfill. At the same time, including only AI agents also limits the empirical insights from experiments. For example, researching fairness norms through formal games without human involvement implicitly assumes that MAAI systems behave accordingly in practical applications (Deck et al., 2026). While it is uncertain whether this assumption holds for MAAI systems, research has shown that human behavior in the same formal games can differ significantly depending on context (Khadjavi & Lange, 2013). Crucially, observing the strategies employed by the agents also does not reveal why they behave the way they do. Selecting irrational strategies—judged from a payoff-maximizing perspective—can happen due to limited intelligence, adherence to a fairness norm, or other unknown reasons.

Leaving humans out of the question raises novel philosophical questions about the meaning of and reasoning about fairness among autonomous agents. We conjecture, however, that to meaningfully evaluate and achieve fairness in MAAI from a human-centric perspective, humans must be integrated into the research design with specific roles. This includes users and auditors who are responsible for fairness outcomes (*Fairness with Human Oversight*) and affected parties who are subject to these outcomes (*Perceived Fairness, Empowering Fairness*). The need is even acknowledged by authors of studies within the *Petri Dish Fairness* archetype (Piatti et al., 2024). We expect that the methodological choice to leave out humans in the included papers is due to the convenience of using AI agents over human subjects. To understand how fairness unfolds in realistic environments where humans may be part of all aspects of a workflow, future work should dedicate efforts to empirical lab and field experiments with human involvement that reflect real-world use contexts.

## 7    Conclusion

This scoping literature review is the first to synthesize the novel and rapidly growing body of research on fairness in MAAI systems. Through a human-centric and systemic lens, existing research is clustered across five dimensions: Fairness Concept, Method, Inception, Objective, and Autonomy. From these clusters, we derive five archetypes characterizing the current state of the field: Normative Delegation, Fairness Effectiveness, Fairness Facade, Fairness Schooling, and Petri Dish Fairness.

Normative Delegation refers to the phenomenon of shifting the responsibility towards the creators of the underlying foundational model by assuming that the foundation models possess a normatively sound, unambiguous conception of fairness. We argue that this assumption is tenuous and empirically unsupported. Instead, fairness requires normative reasoning and specificity on the part of researchers and system designers. Fairness Effectiveness characterizes papers that make fairness specific and measurable, e.g., using metrics or experimental interventions. For empirical insights about fairness in MAAI systems, future research should be guided by these approaches and explore ways to establish reliable fairness measures. Fairness Facade describes the common tendency to treat fairness superficially by implementing fairness through shallow, high-level means such as prompting AI agents to "be fair". Also, the current focus on post-hoc fixes or prompting-based fairness reflects a development–deployment gap that limits robustness and ethical depth. Instead, Fairness Schooling describes the implementation of specific fairness conceptions in the development stages of MAAI systems. While this shift from post-hoc considerations to integral design requirements may improve the robustness, research is still scarce and requires novel methodological approaches to embed fairness early on in the development process. Finally, Petri Dish Fairness expresses the most substantial limitation of current research, with almost all surveyed papers lacking human integration in "fair" MAAI systems. From a human-centric perspective, insights and impact of research for fairness in MAAI systems run void when human stakeholders are absent. Instead, fairness considerations should not only be included technically, but also in the form of human participation.

### 7.1    Limitations

A key limitation of our work is that our scoping review only provides a snapshot of an emergent and fast-moving field with limited validity and generalizability. First, the majority of our literature corpus is from the last two years and only published on the non-peer-reviewed arXiv database, which means that the methodological quality cannot be ensured. Second, many of the included papers are not predominantly concerned with fairness-centered design or evaluation but rather preliminary studies where fairness is a secondary concern. Third, our work does not include publications outside of the academic context, such as industry reports, and, due to the rapidly evolving field, new relevant studies might have been published during the writing process of this paper. Lastly, although we provide counts of papers for each archetype, our work is a purely qualitative analysis and is not to be interpreted as quantitative evidence. Given the comparably small size of papers, some findings presented in this are based on a small number of studies and should therefore be viewed as preliminary rather than definitive.

## 7.2 Outlook

Future research should follow a human-centric and systemic level perspective, moving beyond superficial fixes toward a robust, normatively grounded conception of fairness. This requires embedding fairness considerations early in the system's development rather than relying on post-hoc band aids or "ethical" labels Following the archetype of Fairness Schooling, MAAI systems should have fairness considerations at their core, enabled by techniques such as LLM finetuning. Evaluations should move beyond formal confined settings, towards artefacts resembling real-world use cases, involving the human stakeholders, central to the human-centric paradigm. Research gaps like language variation and underlying foundational models should be addressed. Only then can MAAI systems meet the demands of the fairness desiderata (Deck, Schomäcker, et al., 2024) and evolve from convenience-driven artifacts into truly responsible technologies. Only then can MAAI systems meet the demands of the fairness desiderata (Deck, Schomäcker, et al., 2024) and evolve from convenience-driven artifacts into truly responsible technologies.